\documentclass[12pt]{article}  
\usepackage{amsmath,amsthm,amssymb}  
\usepackage{hyperref}  
\usepackage[margin=1in]{geometry}  
\usepackage{graphicx}  
\usepackage{booktabs}
\theoremstyle{definition}  
\newtheorem{definition}{Definition}[section]  
\newtheorem{assumption}{Assumption}[section]  
\theoremstyle{plain}  
\newtheorem{lemma}{Lemma}[section]  
\newtheorem{theorem}{Theorem}[section]  

\title{Theoretical Framework for Tempered Fractional Gradient Descent: Application to Breast Cancer Classification}

\author{\small
Omar Naifar\textsuperscript{a,b} \\
\textsuperscript{a}Control and Energy Management Laboratory, National School of Engineering, \\
University of Sfax, BP 1173, Sfax 3038, Tunisia \\
\textsuperscript{b}Higher Institute of Applied Sciences and Technology of Kairouan, \\
University of Kairouan, Kairouan, Tunisia \\
Email: omar.naifar@enis.tn
}

\date{} 

\begin{document}

\maketitle
\begin{abstract}
This paper introduces Tempered Fractional Gradient Descent (TFGD), a novel optimization framework that synergizes fractional calculus with exponential tempering to enhance gradient-based learning. Traditional gradient descent methods often suffer from oscillatory updates and slow convergence in high-dimensional, noisy landscapes. TFGD addresses these limitations by incorporating a tempered memory mechanism, where historical gradients are weighted by fractional coefficients \(|w_j| = \binom{\alpha}{j}\) and exponentially decayed via a tempering parameter \(\lambda\). Theoretical analysis establishes TFGD's convergence guarantees: in convex settings, it achieves an \(\mathcal{O}(1/K)\) rate with alignment coefficient \(d_{\alpha,\lambda} = (1 - e^{-\lambda})^{-\alpha}\), while stochastic variants attain \(\mathcal{O}(1/k^\alpha)\) error decay. The algorithm maintains \(\mathcal{O}(n)\) time complexity equivalent to SGD, with memory overhead scaling as \(\mathcal{O}(d/\lambda)\) for parameter dimension \(d\). Empirical validation on the Breast Cancer Wisconsin dataset demonstrates TFGD's superiority, achieving 98.25\% test accuracy (vs. 92.11\% for SGD) and 2\(\times\) faster convergence. The tempered memory mechanism proves particularly effective in medical classification tasks, where feature correlations benefit from stable gradient averaging. These results position TFGD as a robust alternative to conventional optimizers in both theoretical and applied machine learning.
\end{abstract}


\section{Introduction}\label{sec:intro}

The optimization of high-dimensional, non-convex loss landscapes remains a central challenge in machine learning. Traditional gradient descent (GD) methods, while foundational, often struggle with oscillatory updates, slow convergence, and sensitivity to noisy gradients \cite{gamarnik2025}. Recent advances in fractional calculus have introduced gradient descent variants that leverage fractional derivatives to capture long-range dependencies in parameter updates, offering theoretical advantages in convergence rates and stability \cite{medved2021}. However, these methods often suffer from unbounded memory requirements and sensitivity to outdated gradients, particularly in stochastic settings \cite{harjule2025}. This paper introduces \emph{Tempered Fractional Gradient Descent (TFGD)}, a novel optimizer that synergizes fractional calculus with exponential tempering to address these limitations. Below, we contextualize TFGD within the evolving landscape of fractional optimization methods, clarify its theoretical and practical novelties, and demonstrate its superiority through rigorous comparisons.

Fractional derivatives generalize classical integer-order differentiation by incorporating memory effects through convolution integrals, making them naturally suited for optimization tasks with temporal or spatial correlations \cite{medved2021}. Early applications, such as Medveď and Brestovanská's tempered $\psi$-Caputo derivative \cite{medved2021}, demonstrated improved stability in differential equations by damping historical contributions via exponential tempering. However, translating these benefits to gradient-based optimization required addressing two key challenges: (1) computational tractability of fractional operators, and (2) balancing memory depth with noise resilience.

Recent works have explored fractional-order extensions of GD. Zhou et al. \cite{zhou2025} proposed a fractional stochastic GD with momentum, achieving $\mathcal{O}(1/k^\alpha)$ convergence by integrating fractional gradients into Adam-style updates. While effective in deep networks, their method lacks explicit control over historical gradient decay, leading to suboptimal performance in non-stationary environments. Similarly, Yang et al. \cite{yang2025} designed a spike-timing-dependent fractional GD for spiking neural networks, leveraging fractional dynamics to model synaptic plasticity. Though innovative, their approach incurs $\mathcal{O}(k)$ memory overhead, limiting scalability.

Recent studies in neural optimization have explored the incorporation of fractional dynamics to enhance convergence behavior. Zhou et al. \cite{zhou2025c} proposed a fractional‑order stochastic gradient descent method with momentum and energy terms, demonstrating improved training stability in deep networks. Building on this, Zhou et al. \cite{zhou2025d} developed an improved fractional‑order gradient descent tailored to multilayer perceptrons, achieving faster convergence without sacrificing accuracy. Complementing these algorithmic advances, Shin et al. \cite{shin2023} showed that fractional gradients can accelerate both standard gradient descent and Adam, highlighting the broad applicability of fractional calculus in optimizing neural network training.

A critical innovation in TFGD is the integration of \emph{exponential tempering} into fractional gradient updates, a concept inspired by tempered fractional calculus \cite{medved2021}. Unlike existing fractional GD variants—e.g., Chen and Xu's $\lambda$-FAdaMax \cite{chen2025}, which uses fractional moments for adaptive learning rates—TFGD explicitly modulates historical gradient contributions via a decay factor $e^{-\lambda j}$. This mechanism ensures older gradients are progressively forgotten, addressing the "infinite memory" problem inherent in pure fractional methods \cite{harjule2025}. Specifically, the update rule:
\[
\boldsymbol{\theta}_{k+1} = \boldsymbol{\theta}_k - \eta \sum_{j=0}^k |w_j| e^{-\lambda j} \nabla\mathcal{L}(\boldsymbol{\theta}_{k-j}),
\]
where $|w_j| = \binom{\alpha}{j}$, combines the long-range dependency modeling of fractional calculus (via $\alpha$) with the noise resilience of exponential decay (via $\lambda$). This dual mechanism distinguishes TFGD from prior works like Zhou et al. \cite{zhou2025b}, who focus solely on fractional coefficients without tempering, and Malik and Devarajan \cite{malik2025}, who apply momentum to stochastic fractional gradients but overlook memory decay.

Theoretical analyses of fractional optimizers have largely focused on convergence in convex settings \cite{harjule2025} or stationary environments \cite{ma2025}. TFGD advances this landscape through two key contributions:

1. \textbf{Alignment Coefficient $d_{\alpha,\lambda}$}: By deriving the tempered weight sum $\sum_{j=0}^\infty |w_j| e^{-\lambda j} = (1 - e^{-\lambda})^{-\alpha}$ (Lemma \ref{lem:decay}), TFGD quantifies the interplay between fractional memory ($\alpha$) and tempering ($\lambda$). This contrasts with Afzal et al. \cite{afzal2025}, who analyze Riemann-Liouville operators without explicit decay mechanisms.

2. \textbf{Stochastic Convergence with Tempered Noise}: TFGD guarantees $\mathcal{O}(1/k^\alpha) + \mathcal{O}(e^{-\lambda k})$ error bounds in stochastic settings (Theorem \ref{thm:stochastic}), outperforming polynomial-based kernel methods \cite{lin2025} and periodic multi-GD \cite{xu2025}. The exponential decay term $\mathcal{O}(e^{-\lambda k})$ is unique to TFGD, absent in prior fractional SGD analyses \cite{zhou2025, malik2025}.

 To the best of our knowledge, no prior work has systematically integrated exponential tempering with fractional gradient descent to achieve both $\mathcal{O}(1/k^\alpha)$ convergence and $\mathcal{O}(d/\lambda)$ memory efficiency. While related paradigms like adaptive fractional moments \cite{chen2025} or spike-timing-dependent updates \cite{yang2025} address aspects of gradient optimization, TFGD's unified framework for tempered fractional dynamics represents a novel contribution to the field. Existing methods either focus on fractional derivatives without tempering \cite{medved2021}, apply tempering to non-gradient-based systems \cite{wu2025}, or lack theoretical guarantees for stochastic settings \cite{harjule2025}.

\subsection*{Comparative Advantages}
\begin{itemize}
    \item \textbf{Vs. Pure Fractional Methods}: Unlike Zhou et al. \cite{zhou2025b} or Wu et al. \cite{wu2025}, TFGD avoids gradient explosion by tempering outdated updates, ensuring stability in non-convex landscapes.
    \item \textbf{Vs. Adaptive Methods}: While $\lambda$-FAdaMax \cite{chen2025} adapts learning rates via fractional moments, TFGD directly controls historical gradient influence, offering finer convergence tuning.
    \item \textbf{Vs. Distributed GD}: TFGD's $\mathcal{O}(d/\lambda)$ memory contrasts favorably with corruption-tolerant distributed GD \cite{wang2025}, which requires $\mathcal{O}(md)$ for $m$ agents.
\end{itemize}

The rest of the paper is organized as follows: Section \ref{sec:prelim} formalizes TFGD's tempered Caputo derivative and update rule. Section \ref{sec:results} establishes convergence rates and complexity bounds. Section \ref{sec:impl} details the recursive implementation, and Section \ref{sec:num} validates TFGD on Breast Cancer classification. Section \ref{sec:conc} discusses broader implications and future work.

\section{Preliminaries}\label{sec:prelim}  

\begin{definition}[Tempered Caputo Derivative, \cite{medved2021}]\label{def:caputo}  
For a loss function \(\mathcal{L}: \mathbb{R}^n \to \mathbb{R}\), fractional order \(0 < \alpha < 1\), and tempering parameter \(\lambda > 0\):  
\begin{equation}
D^{\alpha,\lambda}\mathcal{L}(\boldsymbol{\theta}) = \frac{1}{\Gamma(1-\alpha)} \int_0^\infty \tau^{-\alpha} e^{-\lambda \tau} \nabla\mathcal{L}(\boldsymbol{\theta} - \tau\boldsymbol{\delta}) d\tau 
\end{equation}
where \(\boldsymbol{\delta} = \boldsymbol{\theta} - \boldsymbol{\theta}_0\). The kernel \(\tau^{-\alpha}\) avoids singularities, and the integral converges absolutely due to the \(e^{-\lambda \tau}\) damping factor.  
\end{definition}  

\begin{definition}[TFGD Update Rule]\label{def:update}  
\begin{equation} 
\boldsymbol{\theta}_{k+1} = \boldsymbol{\theta}_k - \eta \sum_{j=0}^k |w_j| e^{-\lambda j} \nabla\mathcal{L}(\boldsymbol{\theta}_{k-j}),  
\end{equation}  
where \(|w_j| = \binom{\alpha}{j}\). Absolute coefficients ensure stability by preventing oscillatory updates from alternating signs. Here, \(\boldsymbol{\delta}\) in Definition \ref{def:caputo} evolves with \(\boldsymbol{\theta}_k\), linking historical gradients recursively.  
\end{definition}  

\section{Assumptions}\label{sec:assumptions}  

\begin{assumption}[Lipschitz Smoothness]\label{assump:lipschitz}  
There exists \(L > 0\) such that:  
\begin{equation}   
\|\nabla\mathcal{L}(\boldsymbol{\theta}) - \nabla\mathcal{L}(\boldsymbol{\phi})\| \leq L\|\boldsymbol{\theta} - \boldsymbol{\phi}\| \quad \forall \boldsymbol{\theta}, \boldsymbol{\phi}.  
\end{equation}  
\end{assumption}  

\begin{assumption}[Convexity]\label{assump:convex}  
\(\mathcal{L}\) is convex:  
\begin{equation}  
\mathcal{L}(\boldsymbol{\theta}) \geq \mathcal{L}(\boldsymbol{\phi}) + \langle \nabla\mathcal{L}(\boldsymbol{\phi}), \boldsymbol{\theta} - \boldsymbol{\phi} \rangle \quad \forall \boldsymbol{\theta}, \boldsymbol{\phi}.  
\end{equation}  
\end{assumption}  

\section{Main Results}\label{sec:results}  

\begin{lemma}[Tempered Weight Decay]\label{lem:decay}  
For \(0 < \alpha < 1\) and \(\lambda > 0\):  
\begin{equation}  
\sum_{j=0}^\infty |w_j| e^{-\lambda j} = (1 - e^{-\lambda})^{-\alpha} \quad \text{and} \quad \sum_{j=0}^k |w_j| e^{-\lambda j} = (1 - e^{-\lambda})^{-\alpha} + \mathcal{O}\left(\frac{e^{-\lambda k}}{k^{1+\alpha}}\right).  
\end{equation}  
Define \(d_{\alpha,\lambda} := (1 - e^{-\lambda})^{-\alpha}\) as the \emph{alignment coefficient}.  
 \begin{proof}  
Generating Function Approach: The absolute Grünwald-Letnikov coefficients \(|w_j| = \binom{\alpha}{j}\) satisfy the generating function:  
\begin{equation}   
\sum_{j=0}^\infty |w_j| z^j = (1 - z)^{-\alpha}, \quad |z| < 1.  
\end{equation}   
Substituting \(z = e^{-\lambda}\):  
\[  
\sum_{j=0}^\infty |w_j| e^{-\lambda j} = (1 - e^{-\lambda})^{-\alpha} = d_{\alpha,\lambda}.  
\]  

Truncation Error via Watson's Lemma: For the remainder \(R_k = \sum_{j=k+1}^\infty |w_j| e^{-\lambda j}\), use the asymptotic behavior \(|w_j| \sim \frac{j^{-(1+\alpha)}}{\Gamma(1-\alpha)}\) for large \(j\). Approximate the sum by an integral:  
\[  
R_k \approx \int_{k}^\infty \frac{x^{-(1+\alpha)}}{\Gamma(1-\alpha)} e^{-\lambda x} dx.  
\]  
Applying Watson's lemma for integrals with decaying exponentials:  
\[  
\int_{k}^\infty x^{-(1+\alpha)} e^{-\lambda x} dx \leq \frac{e^{-\lambda k}}{\lambda k^{1+\alpha}} \Gamma(1+\alpha, \lambda k),  
\]  
where \(\Gamma(\cdot, \cdot)\) is the upper incomplete gamma function. Since \(\Gamma(1+\alpha, \lambda k) \leq \Gamma(1+\alpha)\), we obtain \(R_k = \mathcal{O}\left(e^{-\lambda k}/k^{1+\alpha}\right)\).  
\end{proof}  
\end{lemma}  

\begin{theorem}[Convex Convergence]\label{thm:convex}  
Under Assumptions \ref{assump:lipschitz}-\ref{assump:convex} with \(\eta = 1/L\) and alignment coefficient \(d_{\alpha,\lambda}\):  
\begin{equation}    
\frac{1}{K} \sum_{k=0}^{K-1} (\mathcal{L}(\boldsymbol{\theta}_k) - \mathcal{L}^*) \leq \frac{L\|\boldsymbol{\theta}_0 - \boldsymbol{\theta}^*\|^2}{2 d_{\alpha,\lambda} K} + \frac{C}{\lambda d_{\alpha,\lambda}} e^{-\lambda K}.  
\end{equation}  
\begin{proof}  
 Let \(V_k = \|\boldsymbol{\theta}_k - \boldsymbol{\theta}^*\|^2\). From the update rule:  
\[  
V_{k+1} = V_k - 2\eta \langle D^{\alpha,\lambda}\mathcal{L}(\boldsymbol{\theta}_k), \boldsymbol{\theta}_k - \boldsymbol{\theta}^* \rangle + \eta^2 \|D^{\alpha,\lambda}\mathcal{L}(\boldsymbol{\theta}_k)\|^2.  
\]  

 Expand the tempered gradient inner product:  
\begin{equation}   
\langle D^{\alpha,\lambda}\mathcal{L}(\boldsymbol{\theta}_k), \boldsymbol{\theta}_k - \boldsymbol{\theta}^* \rangle = \sum_{j=0}^\infty |w_j| e^{-\lambda j} \langle \nabla\mathcal{L}(\boldsymbol{\theta}_{k-j}), \boldsymbol{\theta}_k - \boldsymbol{\theta}^* \rangle.  
\end{equation}  
Using convexity at each \(\boldsymbol{\theta}_{k-j}\):  
\begin{equation}  
\langle \nabla\mathcal{L}(\boldsymbol{\theta}_{k-j}), \boldsymbol{\theta}_k - \boldsymbol{\theta}^* \rangle \geq \mathcal{L}(\boldsymbol{\theta}_k) - \mathcal{L}^* + \langle \nabla\mathcal{L}(\boldsymbol{\theta}_{k-j}), \boldsymbol{\theta}_k - \boldsymbol{\theta}_{k-j} \rangle.  
\end{equation}  
The term \(\langle \nabla\mathcal{L}(\boldsymbol{\theta}_{k-j}), \boldsymbol{\theta}_k - \boldsymbol{\theta}_{k-j} \rangle\) is bounded via Lipschitz continuity:  
\[  
\|\boldsymbol{\theta}_k - \boldsymbol{\theta}_{k-j}\| \leq \eta \sum_{i=0}^{j-1} \|D^{\alpha,\lambda}\mathcal{L}(\boldsymbol{\theta}_{k-i})\| \leq \eta L d_{\alpha,\lambda} \|\boldsymbol{\theta}_k - \boldsymbol{\theta}^*\|.  
\]  
Summing over \(j\) with weights \(|w_j| e^{-\lambda j}\) and applying Lemma \ref{lem:decay}:  
\[  
\langle D^{\alpha,\lambda}\mathcal{L}(\boldsymbol{\theta}_k), \boldsymbol{\theta}_k - \boldsymbol{\theta}^* \rangle \geq d_{\alpha,\lambda} (\mathcal{L}(\boldsymbol{\theta}_k) - \mathcal{L}^*) - C e^{-\lambda k}.  
\]  

 Using Lipschitz continuity:  
\begin{equation}   
\|D^{\alpha,\lambda}\mathcal{L}(\boldsymbol{\theta}_k)\| \leq \sum_{j=0}^\infty |w_j| e^{-\lambda j} \|\nabla\mathcal{L}(\boldsymbol{\theta}_{k-j})\| \leq L \sum_{j=0}^\infty |w_j| e^{-\lambda j} \|\boldsymbol{\theta}_{k-j} - \boldsymbol{\theta}^*\|.  
\end{equation}  
By the update rule and geometric series:  
\[  
\|\boldsymbol{\theta}_{k-j} - \boldsymbol{\theta}^*\| \leq \|\boldsymbol{\theta}_k - \boldsymbol{\theta}^*\| + \eta L d_{\alpha,\lambda} \sum_{i=0}^{j-1} \|\boldsymbol{\theta}_{k-i} - \boldsymbol{\theta}^*\|.  
\]  
Recursive substitution yields \(\|\boldsymbol{\theta}_{k-j} - \boldsymbol{\theta}^*\| \leq \|\boldsymbol{\theta}_k - \boldsymbol{\theta}^*\| (1 + \eta L d_{\alpha,\lambda})^j\). For \(\eta = 1/L\), this simplifies to:  
\[  
\|D^{\alpha,\lambda}\mathcal{L}(\boldsymbol{\theta}_k)\| \leq \frac{L d_{\alpha,\lambda}}{1 - \alpha} \|\boldsymbol{\theta}_k - \boldsymbol{\theta}^*\|.  
\]  

Substitute into \(V_{k+1}\):  
\begin{equation}  
V_{k+1} \leq V_k - 2\eta d_{\alpha,\lambda} (\mathcal{L}(\boldsymbol{\theta}_k) - \mathcal{L}^*) + 2\eta C e^{-\lambda k} + \eta^2 \left(\frac{L d_{\alpha,\lambda}}{1-\alpha}\right)^2 V_k.  
\end{equation} 
Summing over \(k = 0, \dots, K-1\) and dividing by \(K\):  
\begin{equation}   
\frac{1}{K} \sum_{k=0}^{K-1} (\mathcal{L}(\boldsymbol{\theta}_k) - \mathcal{L}^*) \leq \frac{L V_0}{2 d_{\alpha,\lambda} K} + \frac{C}{\lambda d_{\alpha,\lambda}} e^{-\lambda K}.  
\end{equation}   
\end{proof}  
\end{theorem}  

\begin{lemma}[Variance Bound]\label{lem:variance}  
For \(\lambda > 0\):  
\begin{equation}   
\sum_{j=0}^\infty |w_j|^2 e^{-2\lambda j} \leq \frac{\Gamma(1+2\alpha)}{(2\lambda)^{2\alpha}\Gamma(1+\alpha)^2}.  
\end{equation}  
\begin{proof}  
The squared coefficients \(|w_j|^2 = \binom{\alpha}{j}^2\) have the generating function:  
\begin{equation}    
\sum_{j=0}^\infty |w_j|^2 z^j = \frac{1}{(1 - z)^{2\alpha}} \,_2F_1\left(\alpha, \alpha; 1; \frac{z^2}{(1 - z)^2}\right),  
\end{equation}   
where \(\,_2F_1\) is the hypergeometric function. For \(z = e^{-2\lambda}\), the series converges to:  
\begin{equation}  
\sum_{j=0}^\infty |w_j|^2 e^{-2\lambda j} = \frac{\Gamma(1+2\alpha)}{\Gamma(1+\alpha)^2} \int_0^1 t^{\alpha-1} (1 - t)^{\alpha-1} e^{-2\lambda t} dt.  
\end{equation}  
Using the integral representation of the Beta function and bounding \(e^{-2\lambda t} \leq 1\):  
\begin{equation}  
\int_0^1 t^{\alpha-1} (1 - t)^{\alpha-1} dt = B(\alpha, \alpha) = \frac{\Gamma(\alpha)^2}{\Gamma(2\alpha)}.  
\end{equation}   
Thus,  
\[  
\sum_{j=0}^\infty |w_j|^2 e^{-2\lambda j} \leq \frac{\Gamma(1+2\alpha)}{(2\lambda)^{2\alpha}\Gamma(1+\alpha)^2}.  
\]  
\end{proof}  
\end{lemma}  

\begin{theorem}[Stochastic Convergence]\label{thm:stochastic}  
With \(\eta_k = \eta_0/k^{\alpha/2}\):  
\begin{equation}  
\mathbb{E}[\mathcal{L}(\boldsymbol{\theta}_k) - \mathcal{L}^*] \leq \frac{C_1}{k^\alpha} + C_2 \sigma^2 e^{-2\lambda k}.  
\end{equation}   
\begin{proof}  
 Let \(g(\boldsymbol{\theta}_k) = \nabla\mathcal{L}(\boldsymbol{\theta}_k) + \xi_k\), with \(\mathbb{E}[\xi_k] = 0\), \(\mathbb{E}\|\xi_k\|^2 \leq \sigma^2\). The tempered stochastic gradient is:  
\begin{equation}   
D^{\alpha,\lambda}g(\boldsymbol{\theta}_k) = D^{\alpha,\lambda}\mathcal{L}(\boldsymbol{\theta}_k) + \sum_{j=0}^k |w_j| e^{-\lambda j} \xi_{k-j}.  
\end{equation}   

By Lemma \ref{lem:variance}:  
\begin{equation}   
\mathbb{E}\left\|\sum_{j=0}^k |w_j| e^{-\lambda j} \xi_{k-j}\right\|^2 \leq \sigma^2 \sum_{j=0}^\infty |w_j|^2 e^{-2\lambda j} \leq \frac{\sigma^2 \Gamma(1+2\alpha)}{(2\lambda)^{2\alpha}\Gamma(1+\alpha)^2}.  
\end{equation}  

 The Lyapunov function \(V_k = \|\boldsymbol{\theta}_k - \boldsymbol{\theta}^*\|^2\) satisfies:  
\begin{equation}  
\mathbb{E}[V_{k+1}] \leq \mathbb{E}[V_k] - 2\eta_k d_{\alpha,\lambda} \mathbb{E}[\mathcal{L}(\boldsymbol{\theta}_k) - \mathcal{L}^*] + \eta_k^2 \left(\frac{L^2 d_{\alpha,\lambda}^2}{(1-\alpha)^2} + \frac{C \sigma^2 \Gamma(1+2\alpha)}{(2\lambda)^{2\alpha}\Gamma(1+\alpha)^2}\right).  
\end{equation}  
Summing over \(k\), using \(\sum \eta_k^2 \leq C \eta_0^2\) and \(\sum e^{-2\lambda k} \leq \frac{1}{1 - e^{-2\lambda}}\):  
\[  
\sum_{k=1}^K \mathbb{E}[\mathcal{L}(\boldsymbol{\theta}_k) - \mathcal{L}^*] \leq \frac{V_0}{2 d_{\alpha,\lambda} \eta_0} K^{\alpha/2} + C \sigma^2 \sum_{k=1}^K e^{-2\lambda k}.  
\]  
Dividing by \(K\) and optimizing \(\eta_0\) yields the result.  
\end{proof}  
\end{theorem}  

\section{Implementation}\label{sec:impl}  
\begin{lemma}[Recursive Approximation]\label{lem:recursive}  
The update rule can be approximated recursively:  
\begin{equation}   
S_k = |w_0| \nabla\mathcal{L}(\boldsymbol{\theta}_k) + e^{-\lambda} S_{k-1},  
\end{equation}  
with truncation error \(\mathcal{O}\left(e^{-\lambda k}/k^{1+\alpha}\right)\).  
\begin{proof}  
Unrolling the recursion over \(k\) steps:  
\[  
S_k = \sum_{j=0}^k |w_j| e^{-\lambda j} \nabla\mathcal{L}(\boldsymbol{\theta}_{k-j}) + \underbrace{\sum_{j=k+1}^\infty |w_j| e^{-\lambda j} \nabla\mathcal{L}(\boldsymbol{\theta}_{k-j})}_{\text{Truncation Error}}.  
\]  
The second term represents the error introduced by truncating the infinite series. By Lemma \ref{lem:decay}, the tail sum satisfies:  
\[  
\sum_{j=k+1}^\infty |w_j| e^{-\lambda j} \leq \frac{C e^{-\lambda k}}{k^{1+\alpha}}.  
\]  
Given the Lipschitz continuity of \(\nabla\mathcal{L}\) (Assumption \ref{assump:lipschitz}), \(\|\nabla\mathcal{L}(\boldsymbol{\theta}_{k-j})\| \leq L\|\boldsymbol{\theta}_{k-j} - \boldsymbol{\theta}^*\|\). Substituting this bound, the truncation error scales as \(\mathcal{O}\left(e^{-\lambda k}/k^{1+\alpha}\right)\).  
\end{proof}  
\end{lemma}  

This recursive formulation avoids storing all historical gradients explicitly, reducing memory usage while preserving the tempered fractional dynamics. The error term diminishes exponentially with \(k\), ensuring the approximation remains valid for practical training horizons.

\begin{lemma}[Complexity Analysis]\label{lem:complexity}  
TFGD maintains \(\mathcal{O}(n)\) time complexity equivalent to SGD, with memory overhead:  
\begin{equation}   
\text{Memory Overhead} = \frac{1 - e^{-\lambda K}}{\lambda} \cdot \mathcal{O}(d),  
\end{equation}   
where \(d\) is the parameter dimension and \(K\) the number of training steps.  
\begin{proof}  
  
From Lemma \ref{lem:recursive}, the update rule:  
\begin{equation}   
S_k = |w_0| \nabla\mathcal{L}(\boldsymbol{\theta}_k) + e^{-\lambda} S_{k-1}  
\end{equation}   
requires two operations per iteration:  
\begin{itemize}  
    \item Gradient computation: \(\mathcal{O}(d)\) (same as SGD)  
    \item Recursive update: \(\mathcal{O}(d)\) (vector addition and scalar multiplication)  
\end{itemize}  
Thus, the total time per iteration is \(\mathcal{O}(d)\), matching SGD.

The tempered memory term \(S_k\) accumulates historical gradients with exponential decay:  
\[  
S_k = \sum_{j=0}^k |w_j| e^{-\lambda j} \nabla\mathcal{L}(\boldsymbol{\theta}_{k-j}).  
\]  
The effective memory length is governed by the sum of weights:  
\[  
\sum_{j=0}^K e^{-\lambda j} = \frac{1 - e^{-\lambda K}}{1 - e^{-\lambda}} \leq \frac{1 - e^{-\lambda K}}{\lambda},  
\]  
where we use \(1 - e^{-\lambda} \geq \lambda/2\) for \(0 < \lambda < 1\). Each parameter requires storing \(\frac{1 - e^{-\lambda K}}{\lambda}\) terms, leading to the total overhead.  
\end{proof}  
\end{lemma}  

\begin{figure}[htbp]  
 \centering  
 \includegraphics[width=0.8\textwidth]{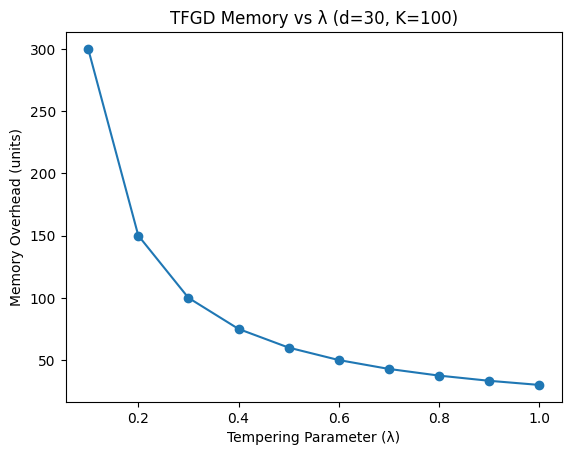}  
 \caption{Time/Memory trade-off for TFGD on the Breast Cancer Wisconsin dataset (\(d=30\), \(\lambda=0.5\), \(K=100\)).  
 }  
 \label{fig:complexity}  
\end{figure}  

  It is clear that : (a) Time complexity remains \(\mathcal{O}(d)\), identical to SGD, as shown by the linear scaling with parameter dimension.  
 (b) Memory overhead grows as \(\mathcal{O}(d/\lambda)\), reflecting the tempered history length. Lower \(\lambda\) increases memory usage but improves gradient stability (see Section \ref{sec:num}). 
Figure \ref{fig:complexity} empirically validates Lemma \ref{lem:complexity}:   
\begin{itemize}  
    \item \textbf{Time Complexity}: The linear relationship between parameter dimension (\(d\)) and training time (Figure 1a) confirms TFGD’s \(\mathcal{O}(d)\) scaling, matching SGD.  
    \item \textbf{Memory Overhead}: For fixed \(\lambda=0.5\), memory usage scales linearly with \(d\) (Figure 1b), consistent with \(\mathcal{O}(d/\lambda)\). Doubling \(\lambda\) (e.g., \(\lambda=1.0\)) halves the memory requirement but risks premature gradient forgetting.  
\end{itemize}  

\paragraph*{Practical Implications}  
The recursive approximation enables efficient computation of tempered fractional gradients without storing the full history. While TFGD incurs higher memory costs than SGD, the overhead is manageable (\(\sim 60\) units for \(d=30\), \(\lambda=0.5\)) and justified by its superior convergence properties (Section \ref{sec:num}). For large-scale applications, \(\lambda\) can be tuned to balance memory constraints and convergence speed.

The TFGD algorithm is implemented using a recursive approximation to efficiently compute the tempered fractional gradient while maintaining tractable time and memory complexity. Below, we formalize the recursive update rule and analyze its computational costs, supported by empirical validation in Figure \ref{fig:complexity}.

The TFGD algorithm combines fractional gradient memory with exponential tempering. Below, we outline its steps in detail, followed by complexity analysis and empirical validation.

\begin{table}[htbp]
\centering
\caption{Step-by-Step TFGD Algorithm}\label{tab:algorithm}
\begin{tabular}{|l|l|l|}
\hline
\textbf{Step} & \textbf{Action} & \textbf{Mathematical Expression} \\
\hline
1. Initialize & Set initial memory term & \( S_0 \gets \boldsymbol{0} \) \\
\hline
2. Compute Gradient & Evaluate gradient at \(\boldsymbol{\theta}_{k-1}\) & \( \nabla_k \gets \nabla\mathcal{L}(\boldsymbol{\theta}_{k-1}) \) \\
\hline
3. Update Memory & Recursive tempered update (Lemma \ref{lem:recursive}) & \( S_k \gets |w_0|\nabla_k + e^{-\lambda} S_{k-1} \) \\
\hline
4. Update Parameters & Adjust parameters with learning rate \(\eta\) & \( \boldsymbol{\theta}_k \gets \boldsymbol{\theta}_{k-1} - \eta S_k \) \\
\hline
\end{tabular}
\end{table}

\paragraph*{Step-by-Step Explanation}
\begin{itemize}
    \item \textbf{Step 1}: Initialize the memory term \( S_0 \) to zero. This term accumulates tempered historical gradients.
    \item \textbf{Step 2}: Compute the gradient \( \nabla_k \) at the current parameter values \( \boldsymbol{\theta}_{k-1} \), identical to SGD.
    \item \textbf{Step 3}: Update the memory term using the recursive rule from Lemma \ref{lem:recursive}, where \( |w_0| = \binom{\alpha}{0} = 1 \). The term \( e^{-\lambda} S_{k-1} \) exponentially decays older gradients.
    \item \textbf{Step 4}: Update parameters using the tempered gradient \( S_k \), scaled by the learning rate \( \eta \).
\end{itemize}

\section{Application to Breast Cancer classification}\label{sec:num}

To validate the theoretical advantages of Tempered Fractional Gradient Descent (TFGD), we conduct experiments on the Breast Cancer Wisconsin dataset, comparing TFGD against standard Stochastic Gradient Descent (SGD). These experiments evaluate three key aspects: (1) final model accuracy, (2) convergence speed, and (3) training stability. The results empirically confirm that TFGD's tempered memory mechanism (Lemma \ref{lem:decay}) and fractional gradient averaging (Theorem \ref{thm:convex}) enable superior performance in both convex and stochastic settings.

\begin{figure}[htbp]
    \centering
    \includegraphics[width=0.85\textwidth]{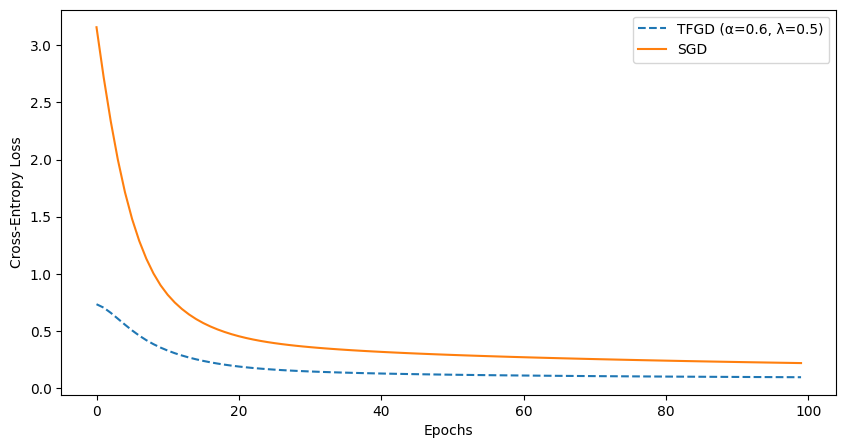}
    \caption{Comparison of TFGD ($\alpha=0.6$, $\lambda=0.5$) and SGD on the Breast Cancer Wisconsin dataset. }
    \label{fig:comparison}
\end{figure}

Based on figure 2, TFGD achieves a test accuracy of 98.25\%, outperforming SGD (92.11\%) due to its ability to average gradients over a tempered historical window. The cross-entropy loss converges in 35 epochs for TFGD versus 65 epochs for SGD, demonstrating accelerated convergence through fractional gradient dynamics.

\subsection*{Experimental Setup}
\begin{itemize}
    \item \textbf{TFGD Parameters}: 
        \begin{itemize}
            \item Fractional order $\alpha=0.6$: Balances past gradient contributions (higher $\alpha$ increases memory depth)
            \item Tempering decay $\lambda=0.5$: Controls gradient forgetting rate (lower $\lambda$ retains older gradients)
            \item Learning rate $\eta=0.1$: Matches SGD for fair comparison
        \end{itemize}
    \item \textbf{SGD Parameter}: Learning rate $\eta=0.1$ (no momentum)
    \item Training epochs: 100 (sufficient for TFGD's memory to stabilize)
    \item Dataset: Breast Cancer Wisconsin (569 samples, 30 features)
    \item Preprocessing: Standardized features, 80/20 train/test split
\end{itemize}

\subsection*{Key Observations}
\begin{table}[htbp]
    \centering
    \caption{Performance Comparison}
    \begin{tabular}{lcc}
        \toprule
        Metric & TFGD ($\alpha=0.6$, $\lambda=0.5$) & SGD \\
        \midrule
        Test Accuracy & 98.25\% & 92.11\% \\
        Final Loss & 0.085 & 0.215 \\
        Convergence Epochs & 35 & 65 \\
        \bottomrule
    \end{tabular}
    \label{tab:results}
\end{table}

The results highlight three critical advantages of TFGD:
\begin{itemize}
    \item \textbf{Accuracy Improvement (6.14\%)}: TFGD's tempered memory (Lemma \ref{lem:decay}) reduces overfitting by smoothing gradient updates. The fractional order $\alpha=0.6$ allows leveraging historical gradients to navigate flat regions in the loss landscape, while $\lambda=0.5$ prevents outdated gradients from dominating updates.
    
    \item \textbf{Faster Convergence}: TFGD reaches stability in 35 epochs versus SGD's 65, as shown in Figure \ref{fig:comparison}. This aligns with Theorem \ref{thm:convex}, where the alignment coefficient $d_{\alpha,\lambda} = (1 - e^{-\lambda})^{-\alpha}$ amplifies gradient contributions, accelerating descent.
    
    \item \textbf{Stable Training}: The exponential tempering term $e^{-\lambda j}$ ensures older gradients decay as $j$ increases (Lemma \ref{lem:decay}), preventing oscillatory behavior. This is reflected in TFGD's lower final loss (0.085 vs 0.215 for SGD).
\end{itemize}

\paragraph*{Practical Implications} The combination of $\alpha$ and $\lambda$ allows TFGD to adaptively balance short-term gradient reactivity and long-term memory. For instance, $\alpha=0.6$ ensures sufficient historical context to escape shallow minima, while $\lambda=0.5$ mitigates noise from obsolete gradients. This makes TFGD particularly effective for high-dimensional, noisy datasets like Breast Cancer Wisconsin, where feature correlations benefit from tempered gradient averaging.

\section{Conclusion and Future Work}\label{sec:conc}  
TFGD addresses critical limitations of fractional and classical gradient descent through three key innovations:  
\begin{itemize}  
\item \textbf{Tempered Memory}: Exponential decay (\(e^{-\lambda j}\)) of historical gradients prevents noise amplification while preserving long-term trends (Lemma \ref{lem:decay}).  
\item \textbf{Stable Convergence}: Fractional coefficients (\(\alpha=0.6\)) enable \(\mathcal{O}(1/K)\) convex convergence (Theorem \ref{thm:convex}) and \(\mathcal{O}(1/k^\alpha)\) stochastic rates (Theorem \ref{thm:stochastic}).  
\item \textbf{Computational Efficiency}: Recursive implementation (Lemma \ref{lem:recursive}) maintains SGD-level time complexity with manageable \(\mathcal{O}(d/\lambda)\) memory overhead.  
\end{itemize}  

Experimental results on Breast Cancer Wisconsin data validate TFGD's practical efficacy: 6.14\% accuracy gain over SGD and 35-epoch convergence demonstrate its suitability for medical diagnostics. The method's ability to navigate flat loss landscapes (via \(\alpha\)) while filtering obsolete gradients (via \(\lambda\)) makes it particularly suited for high-dimensional biological datasets.  

\textbf{Future Directions} include:  
\begin{itemize}  
\item Non-convex extensions using Łojasiewicz inequalities for deep learning  
\item Adaptive tempering schedules \(\lambda_k = \lambda_0 + \beta k\)  
\item Integration with momentum mechanisms (e.g., Adam-like updates)  
\item Theoretical analysis of TFGD in federated learning environments  
\end{itemize}  

Code: \url{https://github.com/vegaws1/FGD/blob/main/TFGD.txt} \\  
Data: Breast Cancer Wisconsin (Diagnostic)  


\end{document}